\documentclass[sigconf]{acmart}
\AtBeginDocument{%
  }

\usepackage{cleveref}
\crefname{section}{Section}{Sections}
\crefname{figure}{Figure}{Figures}
\crefname{table}{Table}{Tables}

\copyrightyear{2026}
\acmYear{2026}
\setcopyright{cc}
\setcctype{by}
\acmConference[MM '26] {Proceedings of the 34th ACM International Conference on Multimedia}{November 10--14, 2026}{Rio de Janeiro, Brazil.}
\acmBooktitle{Proceedings of the 34th ACM International Conference on Multimedia (MM '26), November 10--14, 2026, Rio de Janeiro, Brazil}
\acmISBN{979-8-4007-2213-4/2026/11}
\acmDOI{10.1145/3767308.3835016}

\begin{document}

\title[LASA: Language-and-Source-Anchored Alignment]{LASA: Language-and-Source-Anchored Alignment for Domain Generalized Semantic Segmentation}
\author{Jinhong Zhu}
\orcid{0009-0007-8468-8162}
\affiliation{%
  \institution{Xiamen University}
  \department{Key Laboratory of Multimedia
  Trusted Perception and Efficient
  Computing, Ministry of Education of China}
  \city{Xiamen}
  \state{Fujian}
  \country{China}
}
\email{949278569@qq.com}

\author{Weiqi Yan}
\orcid{0009-0005-5606-1675}
\affiliation{%
  \institution{Xiamen University}
  \department{Key Laboratory of Multimedia
  Trusted Perception and Efficient
  Computing, Ministry of Education of China}
  \city{Xiamen}
  \state{Fujian}
  \country{China}
}
\email{weiqi\_yan@outlook.com}

\author{Shengchuan Zhang}
\orcid{0000-0002-0800-0609}
\correspondingauthor
\affiliation{%
  \institution{Xiamen University}
  \department{Key Laboratory of Multimedia
  Trusted Perception and Efficient
  Computing, Ministry of Education of China}
  \city{Xiamen}
  \state{Fujian}
  \country{China}
}
\email{zsc\_2016@xmu.edu.cn}

\author{Liujuan Cao}
\orcid{0000-0002-7645-9606}
\affiliation{%
  \institution{Xiamen University}
  \department{Key Laboratory of Multimedia
  Trusted Perception and Efficient
  Computing, Ministry of Education of China}
  \city{Xiamen}
  \state{Fujian}
  \country{China}
}
\email{caoliujuan@xmu.edu.cn}



\renewcommand{\shortauthors}{Jinhong Zhu, Weiqi Yan, Shengchuan Zhang \& Liujuan Cao}

\begin{abstract}
    Domain Generalization Semantic Segmentation (DGSS) focuses on generalizing knowledge from labeled source domains to unseen target domains where data is unavailable during the training phase. 
    While conventional methods utilize style randomization or feature normalization to mitigate domain shifts, they often impair feature integrity. 
    Specifically, style randomization distorts the underlying feature manifold due to its coarse-grained nature, while feature normalization suppresses discriminative, domain-sensitive semantic details owing to its rigid design.
    To address these limitations, we propose the \textbf{L}anguage-\textbf{a}nd-\textbf{S}ource-Anchored \textbf{A}lignment \textbf{(LASA)} framework, which comprises three synergistic components: Text-and-Source-Guided Style Transfer (TSGST), Domain-Aware Query Adapter (DAQA), and Domain-Aware Decoder Optimizer (DADO).
    Concretely, the TSGST module addresses manifold distortion by utilizing source features as structural anchors and vision-language model (VLM) priors as fine-grained guidance.
    To restore suppressed discriminative and domain-sensitive details, the DAQA module recalibrates object queries via categorical guidance and domain-aware signatures, while the DADO module aligns the resulting query distributions with a shared classifier to ensure consistent categorical responses across domains.
    Extensive experiments on challenging benchmarks demonstrate that our method significantly outperforms state-of-the-art approaches. 
\end{abstract}




\maketitle

\section{Introduction}
\begin{figure}
    \includegraphics[width=\linewidth]{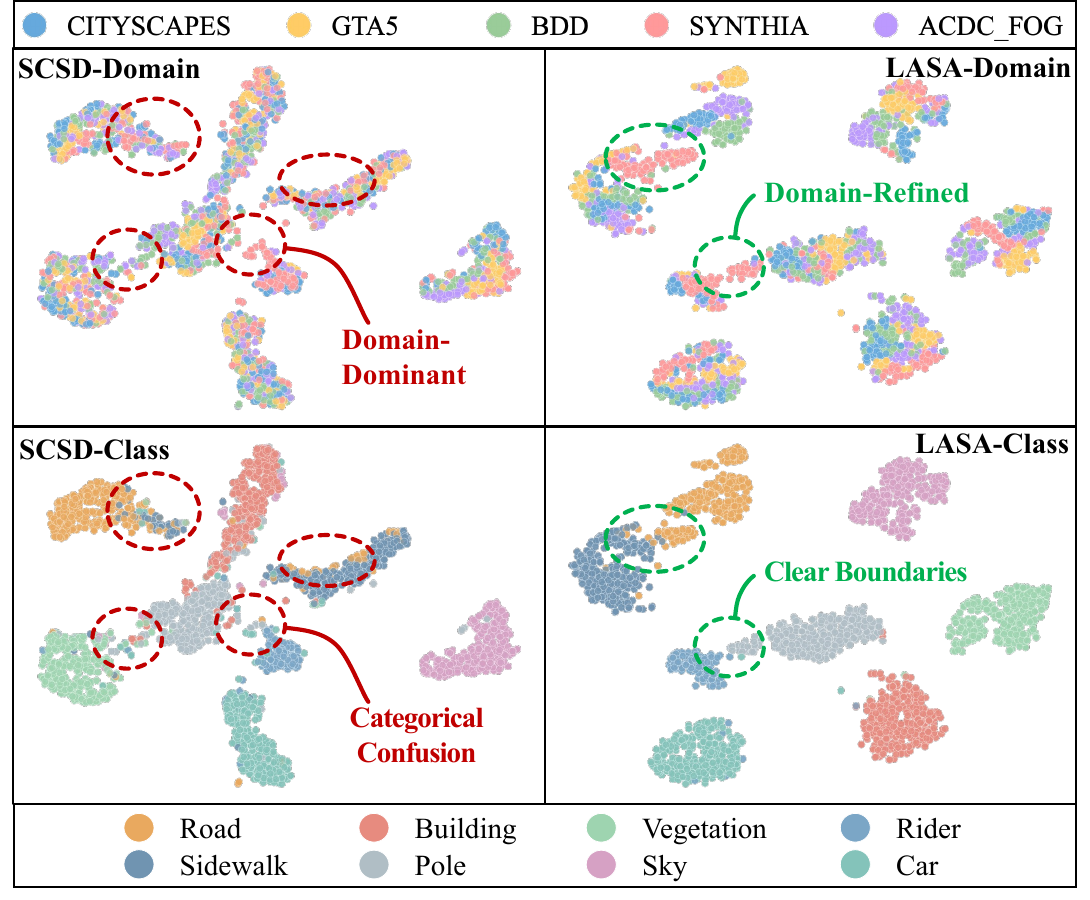}
    \caption[]{T-SNE visualization of image features for the SCSD method (left)~\cite{Niu2025Exploring} and our LASA method (right). In the top row, diverse colors distinguish different domains. In the bottom row, distinct colors represent different classes.}
    \label{fig:feature_space}
\end{figure}

Semantic segmentation~\cite{Long2015Fully,He2017Mask,Cheng2022Masked} serves as a cornerstone of computer vision, providing critical scene understanding for high-stakes applications such as autonomous driving, robotic perception, and medical imaging.
While traditional supervised methods have ach\-ieved remarkable success, their performance heavily relies on the Independent and Identically Distributed (i.i.d.) assumption.
In real-world scenarios, however, models often encounter a ``domain gap''—a distribution shift between the source (training) and target (deployment) domains caused by varying weather, lighting, or geographic locations.
To address this, Domain Generalization Semantic Segmentation (DGSS) has emerged, aiming to learn robust representations from source domains that can generalize to unseen environments without any prior access to target-domain data.
\begin{figure}
    \includegraphics[width=\linewidth]{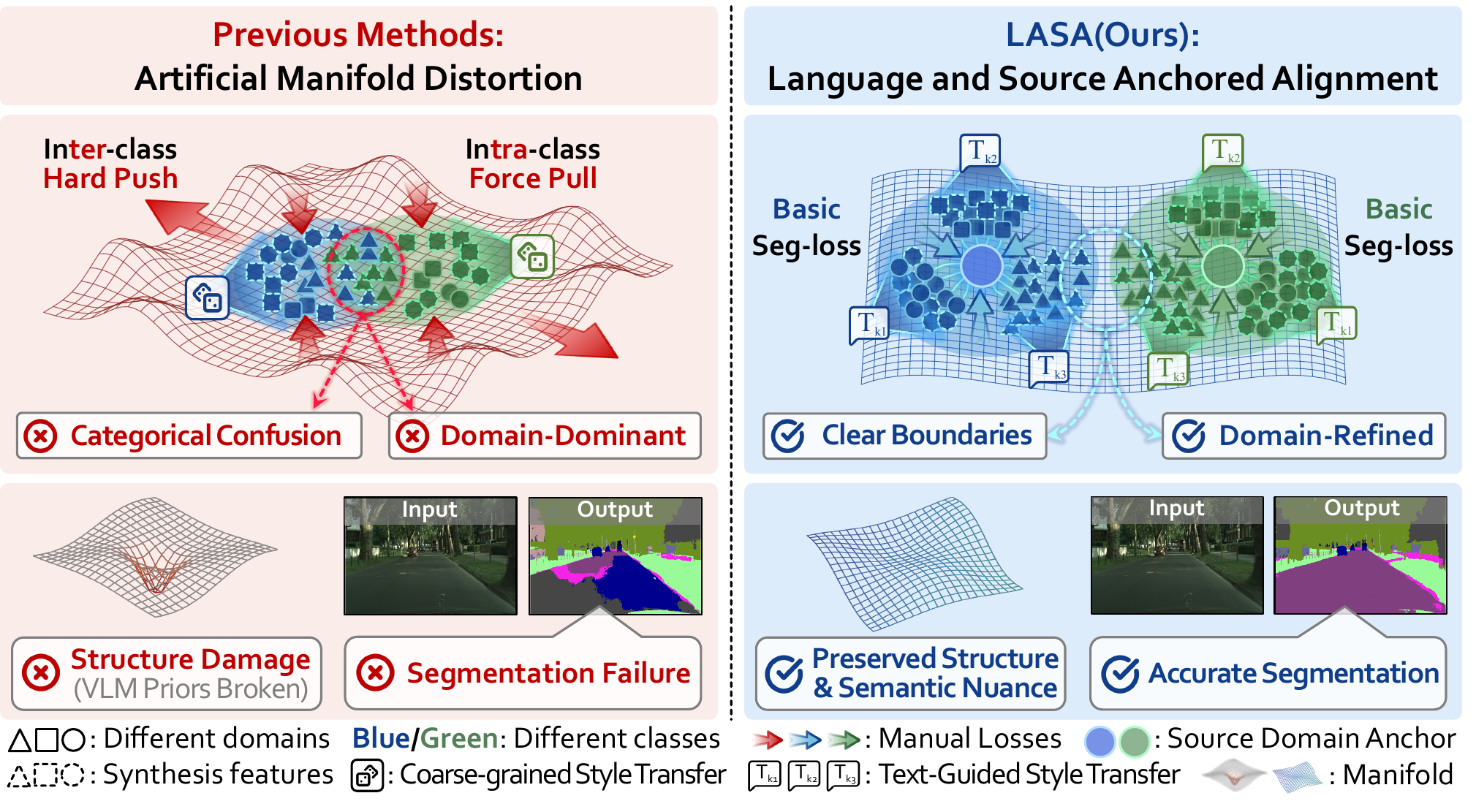}
    \caption{Conceptual illustration of LASA’s motivation. (Left) Prior Methods: Coarse-grained style randomization and rigid feature normalization strategies distort the feature manifold. (Right) LASA (Ours): Anchored by source features and guided by VLM priors, LASA achieves consistent domain alignment.}
    \label{fig:motivation}
\end{figure}
Existing DGSS approaches primarily fall into two categories: Style Randomization and Feature Normalization.
Style randomization methods synthesize diverse samples at the image or feature level to prevent the model from over-fitting the source distribution.
Feature normalization methods employ techniques such as selective whitening to suppress domain-sensitive elements and retain domain-invariant content. Despite their progress, these methods suffer from inherent limitations.
As illustrated in \cref{fig:motivation}, style randomization distorts the underlying feature manifold due to its coarse-grained nature, leading to categorical confusion.
Conversely, feature normalization suppresses discriminative and domain-sensitive semantic details owing to its rigid design, which is vital for precise segmentation.
To further illustrate these issues, we compare the recent state-of-the-art (SOTA) method SCSD~\cite{Niu2025Exploring} with our proposed Language-and-Source-Anchored Alignment (LASA) framework in \cref{fig:feature_space}.
While both methods achieve intra-class feature clustering, existing strategies suffer from the joint negative effects of coarse-grained style randomization and rigid feature normalization.
Specifically, coarse-grained randomization produces domain-dominant features that are indistinguishable from other categories, leading to severe categorical confusion.
Simultaneously, rigid normalization suppresses discriminative and domain-sensitive details, leaving the model unable to effectively organize these domain-dominant features—they become ``lost'' between clustering by domain and clustering by class.
In contrast, LASA addresses manifold distortion by utilizing source features as structural anchors and VLM priors as fine-grained guidance.
This ensures the preservation of the intra-class domain structure, enabling domain-dominant features to remain proximal across different categories while remaining clearly separable by category.
The superiority of this structural preservation is qualitatively validated in \cref{fig:feature_space}, where LASA exhibits significantly tighter class clusters and clearer decision boundaries compared to the recent SCSD~\cite{Niu2025Exploring} method.

To systematically realize such a well-structured manifold, we introduce three synergistic components within the LASA framework: Text-and-Source Guided Style Transfer (TSGST), Domain-Aware Query Adapter (DAQA), and Domain-Aware Decoder Optimizer (DADO).
Specifically, the TSGST module addresses manifold distortion by treating source features as structural anchors and utilizing VLM priors as fine-grained guidance to facilitate controllable style augmentation.
Within this space, the DAQA module provides fine-grained local guidance by recalibrating object queries through the integration of domain-aware signatures and category-specific semantic priors, replenishing suppressed semantic details.
Furthermore, the DAQA module restores suppressed discriminative and domain-sensitive details by recalibrating object queries via domain-aware semantic contexts. The DADO module then aligns the resulting query distributions with a shared classifier to ensure consistent categorical responses across domains.

We evaluate the proposed LASA framework on standard DGSS benchmarks.
Experimental results demonstrate that our model consistently outperforms existing SOTA methods across diverse unseen scenarios.
Notably, LASA achieves remarkable performance gains, particularly in extreme environments: it improves mIoU by 5.48\% on GTAV $\rightarrow$ BDD, 8.91\% on ACDC-Snow, and 8.64\% on ACDC-Night compared to previous methods.
These results underscore the framework's competitive generalization capability across diverse and challenging scenarios.

Our main contributions are summarized as follows:
\begin{itemize}
    \item We propose the LASA framework for DGSS, and extensive experiments have demonstrated its competitive generalization capability across diverse domains.

    \item To address artificial manifold distortion and categorical confusion, we propose the TSGST module, which stabilizes the underlying feature manifold by utilizing source features as structural anchors and cross-modal semantic priors from VLMs to guide controllable style augmentation.

    \item To recover suppressed semantic details and bridge residual domain gaps, we propose the DAQA and DADO modules, which provide local fine-grained modulation and global distribution optimization by dynamically integrating category-specific and domain-aware contexts.
\end{itemize}

\section{Related Works}

\subsection{Domain Generalized Semantic Segmentation}

Domain Generalized Semantic Segmentation (DGSS) aims to train models that generalize to unseen target domains without target data during training~\cite{Choi2021Robustnet,Huang2021Fsdr,Rafi2024Domain,schwonberg2025domain}.
This fundamentally distinguishes it from Unsupervised Domain Adaptation (UDA)~\cite{Hoyer2022Daformer,Xie2023Sepico}, which uses unlabeled target data for alignment.
While early Multi-Source Domain Generalization~\cite{Li2018Domain,Dou2019Domain} focuses on cross-source alignment, Single-Source Domain Generalization (SSDG)~\cite{Lee2022Wildnet,Huang2023Style,Niu2025Exploring} has become more prominent due to the high cost of data annotation.
Existing SSDG methods primarily follow two paradigms: style randomization, which expands the training distribution to cover potential domain shifts, and feature Normalization, which extracts domain-invariant representations by mitigating domain-sensitive style features.
Recent advances leverage vision foundation models like CLIP to provide large-scale pre-trained priors.
Our proposed Language-and-Source-Anchored Alignment (LASA) framework aligns with this cross-modal paradigm, utilizing language-driven semantic priors to guide style and feature generation in a single-source setting.

\subsection{Style Randomization}

Style randomization effectively prevents overfitting to source-specific visual characteristics by synthesizing stylistically diverse unseen domains.
At the image and frequency levels, early works like WildNet~\cite{Lee2022Wildnet} and AdvStyle~\cite{Zhong2022Adversarial} stylize inputs via external data or adversarial training.
Frequency-based methods, such as FSDR~\cite{Huang2021Fsdr} and PASTA~\cite{Chattopadhyay2023Pasta}, decouple style by manipulating amplitude spectra while preserving structural phase information.
Recently, generative models like DGInStyle~\cite{Jia2024Dginstyle} and CLOUDS~\cite{Benigmim2024Collaborating} have introduced diffusion-based augmentation for photorealistic diversity.
To reduce computation, feature-level methods like SiamDoGe~\cite{Wu2022Siamdoge} randomize latent representations. Recent language-guided methods, such as FAMix~\cite{Fahes2024simple} and SCSD~\cite{Niu2025Exploring}, use CLIP text embeddings to direct feature transformations.
In contrast, LASA uses source features as structural anchors to guide the language-driven style transfer. This anchoring mechanism ensures expressive style diversity while strictly maintaining the structural integrity of the feature manifold.
\subsection{Feature Normalization}

Feature normalization enforces the learning of Domain Invariant Features (DIF) by suppressing domain-specific noise or maintaining semantic consistency. Classical techniques like IBN-Net~\cite{Pan2018Two} and RobustNet~\cite{Choi2021Robustnet} use normalization or selective whitening of covariance matrices to decouple style from content. Modern approaches pivot toward high-order statistical alignment and semantic structure preservation. For instance, BlindNet~\cite{Ahn2024Style} and SRMA~\cite{Jiao2024Semantic} employ covariance matching and multi-level semantic reshuffling to learn consistent representations across views. Furthermore, SPC-Net~\cite{Huang2023Style} uses style memory banks and semantic prototypes for cluster-based classification. While these methods aim to produce robust representations, LASA optimizes this process by introducing domain-aware modulation at the query and decoder levels, providing a safeguard for categorical consistency across domains.

\section{Method}

\begin{figure*}
    \includegraphics[width=\textwidth]{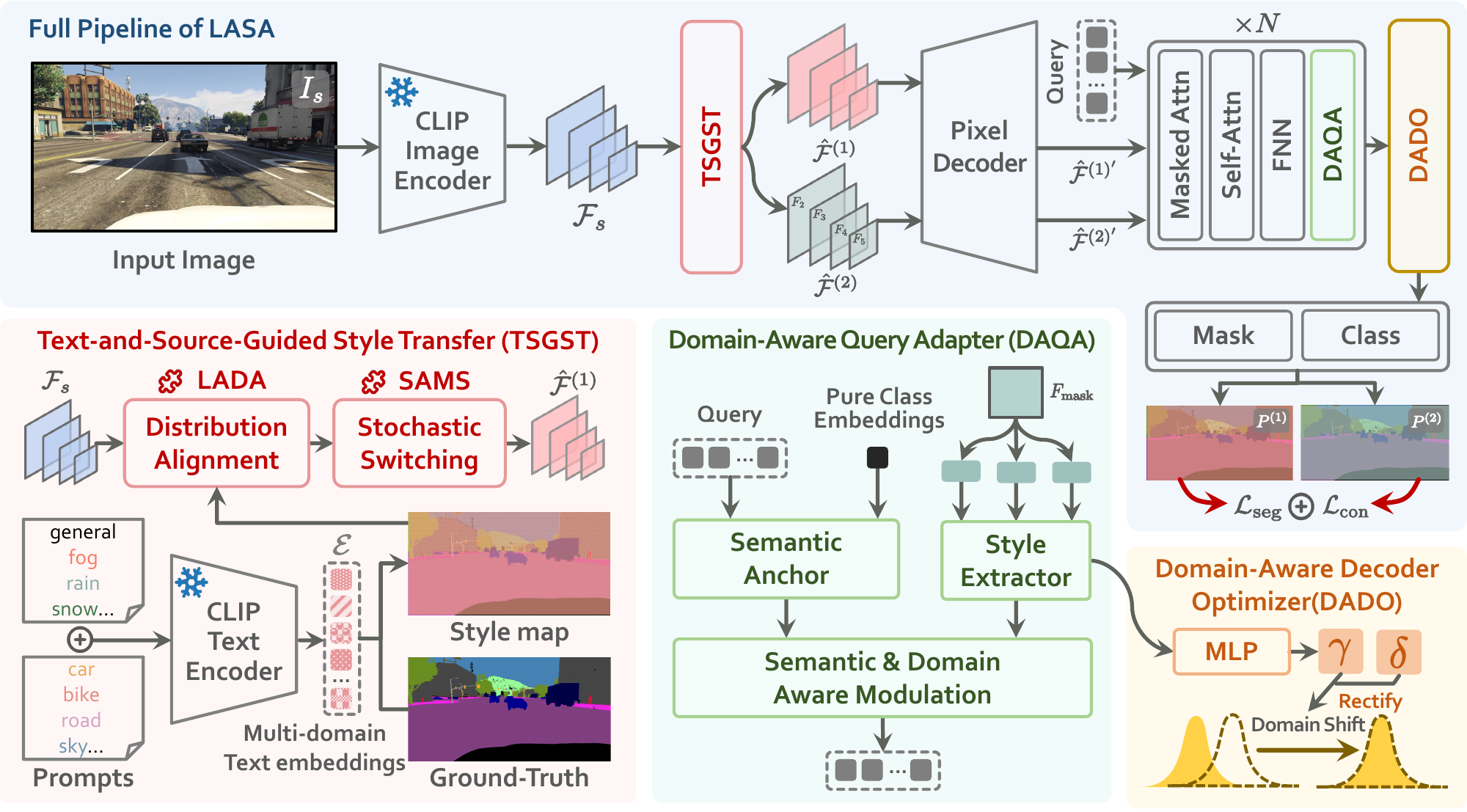}
    \caption{Overview of the proposed LASA, which consists of the TSGST module, the DAQA module, and the DADO module.}
    \label{fig:architecture}
\end{figure*}

\subsection{Overview of LASA}

The architecture of Language-and-Source-Anchored Alignment (LASA) framework is shown in \cref{fig:architecture}, which follows an end-to-end pipeline to ensure robust cross-domain generalization.

Given an input image $I_s$, it is first processed by a frozen pre-trained CLIP visual encoder to obtain the multi-level visual features $\mathcal{F}_s = \{ F_i \in \mathbb{R}^{D \times \frac{H}{2^i} \times \frac{W}{2^i}} \mid i \in \{2,3,4,5\} \}$. These features $\mathcal{F}_s$ are then fed into the Text-and-Source-Guided Style Transfer (TSGST) module, acting as a domain-augmentation engine, to generate two parallel streams of augmented features $\hat{\mathcal{F}}^{(1)}$ and $\hat{\mathcal{F}}^{(2)}$. 

The augmented features are then processed by a pixel decoder to generate dense features $F_{\text{mask}}$, which then interact with object queries $Q$ within the Domain-Aware Query Adapter (DAQA). Acting as an instance-level calibrator, DAQA refines the object queries $Q$ into domain-aware versions $Q^\prime$ by integrating domain signatures with categorical semantic guidance. 

To resolve final distribution disparities, the Domain-Aware Decoder Optimizer (DADO) performs global rectification on the refined queries $Q^\prime$, yielding the final aligned representations $Q^{\prime \prime}$. 

Finally, the framework is optimized by minimizing a joint objective function consisting of a segmentation loss $\mathcal{L}_{\text{seg}}$ for pixel-level accuracy and a consistency loss $\mathcal{L}_{\text{con}}$ between the two branches.

During inference, LASA follows a streamlined single-path execution where the TSGST module is bypassed to ensure efficiency. The visual features $\mathcal{F}$ are processed directly, while the DAQA and DADO modules are retained to perform instance-level recalibration and canonical alignment. This allows the model to leverage the well-structured manifold established during training to achieve robust generalization in unseen target domains.

\subsection{Text-and-Source-Guided Style Transfer}

The Text-and-Source-Guided Style Transfer (TSGST) module is designed to proactively expand diversity of training scenarios while strictly safeguarding the structural integrity of the feature manifold. Rather than performing arbitrary style randomization, TSGST leverages fine-grained vision-language priors and source structural anchors to construct a latent space that is both semantically expressive and geometrically robust.

As illustrated in \cref{fig:architecture}, TSGST processes the hierarchical features $\mathcal{F}_s=\{F_i\}_{i=2}^5$ through a stochastic dual-branch mechanism, generating two parallel feature streams $\hat{\mathcal{F}}^{(1)}$ and $\hat{\mathcal{F}}^{(2)}$ to establish a foundation for consistency learning. TSGST achieves this through three tightly coupled components: Language-Anchored Distribution Alignment (LADA), which steers visual features toward diverse target styles via domain-specific text embeddings; Source-Anchored Manifold Stabilization (SAMS), which stochastically anchors the stylized features to the original source content to prevent representation collapse; and a Dual-Path Consistency Learning scheme, which reconciles cross-branch discrepancies to pull domain-specific variations toward a unified categorical manifold.

\paragraph{Language-Anchored Distribution Alignment.} To translate textual semantics into fine-grained visual guidance, we propose Language-Anchored Distribution Alignment (LADA). The core motivation is to effectively use the joint semantics (category and domain) embedded in textual embeddings to organize the feature space into a structured intra-class inter-domain topology. Let $\mathcal{E}\in \mathbb{R}^{K\times C\times D_t}$ denote the domain-specific textual embeddings, where $K$ and $C$ represent the number of domains and categories, and each $D_t$-dimensional vector encodes both categorical identity and domain context. For each branch $m\in\{1,2\}$, we first randomly sample a domain index $k^{(m)}\in\{1,\ldots,K\}$, which is kept consistent across all hierarchical levels $i\in\{2,3,4,5\}$ within the same branch to ensure stylistic coherence across scales. A crucial component of LADA is the cross-modal projector $\phi(\cdot):\mathbb{R}^{D_t}\rightarrow\mathbb{R}^D$, a learnable mapping that transforms textual embeddings into the visual latent space. Specifically, for each hierarchical feature $F_i\in\mathbb{R}^{\frac{H}{2^i}\times \frac{W}{2^i}\times D}$, we construct a language-anchored style map $M_i^{(m)}\in \mathbb{R}^{\frac{H}{2^i}\times \frac{W}{2^i}\times D}$ by projecting the sampled embedding $E_{k^{(m)},c}$ (e.g., the embedding for "a car in rain" when $c$ is the car category and $k^{(m)}$ is the rain domain), where $c=Y(h,w)$:
\begin{equation}
    M_i^{(m)}(h,w)=\phi(E_{k^{(m)},c}), \quad \text{where } c=Y(h,w)
\end{equation}

To transfer the textual style while maintaining semantic boundaries, LADA performs a channel-wise distribution alignment via a sort-matching operation:
\begin{equation}
    F_{\text{aligned}, i}^{(d)} =
\operatorname{Rank}\!\left(F_i^{(d)}\right)
\circ
\operatorname{Sort}\!\left(M_i^{(d)}\right),
\end{equation}
where $d\in\{1,\ldots,D\}$ is the channel index. Specifically, $\operatorname{Sort}(\cdot)$ extracts the target value set by sorting the elements of $M_i^{(d)}$ in ascending order to define the channel's intensity distribution. Concurrently, $\operatorname{Rank}(\cdot)$ determines the spatial ordinal ranking of pixels in $F_i^{(d)}$ to preserve the spatial topology. The composition operator $\circ$ then assigns the $k$-th smallest value from the sorted target distribution to the exact spatial coordinate of the $k$-th smallest pixel in the source feature. This mechanism ensures that $F_{\text{aligned}, i}$ inherits the statistical characteristics of the category-domain textual priors while remaining strictly anchored to the structural information provided by the source content.

To ensure end-to-end differentiability, we decouple the gradient paths: the gradient flows naturally to the mapping layer $\phi$ through the continuous values of $F_{\text{aligned}, i}$, while for the other trainable modules, we adopt a straight-through estimator (STE) to bypass the non-differentiable sorting operations:
\begin{equation}
    F_{\text{LADA},i}=F_{\text{aligned},i}+F_i-\text{sg}(F_i)
\end{equation}
where $\text{sg}(\cdot)$ denotes the stop-gradient operation.

\paragraph{Source-Anchored Manifold Stabilization.} To preserve the structural stability of the latent space and prevent the mapping $\phi$ from producing distorted or over-fitted representations, we introduce Source-Anchored Manifold Stabilization (SAMS). Since the CLIP backbone is frozen, its original features provide definitive structural anchors. SAMS capitalizes on this by using source features as immutable reference points, providing a critical manifold regularization for the projector $\phi$. This compels the model to generate stylized maps that are not only texturally consistent with the text prompts but also geometrically compatible with the intrinsic visual manifold of CLIP. Specifically, within the TSGST module, SAMS stochastically selects between the original source feature $F_i$ and the stylized feature $F_{\text{LADA},i}$ using a binary selector $s\in\{0,1\}$:
\begin{equation}
    \hat{F}_i =
    \begin{cases}
        F_i, & \text{if } s = 0 \\
        F_{\text{LADA},i}, & \text{if } s = 1
    \end{cases}
\end{equation}
where $s$ follows a Bernoulli distribution with $P(s=0)=0.5$. This ensures that the output features remain anchored to the robust semantic structure of CLIP's original space for half the training instances, preventing representation collapse during style transfer.

\paragraph{Dual-Path Consistency Learning.}
To enforce categorical consistency, we adopt a dual-path optimization scheme generating two parallel multi-level feature sets, $\hat{\mathcal{F}}^{(1)}$ and $\hat{\mathcal{F}}^{(2)}$, from the shared input $\mathcal{F}_s$. By independently sampling the stochastic selectors $s^{(1)}$ and $s^{(2)}$ within their respective TSGST modules, the framework reconciles diverse alignment scenarios, including source-to-augmented and augmented-to-augmented pairs.
The cross-branch discrepancy is regularized via a unidirectional consistency loss:
\begin{equation}
    \mathcal{L}_\text{con}=\text{KL}(\text{sg}(P^{(1)})\parallel P^{(2)})
\end{equation}
where $P^{(m)}$ is the prediction from the $m$-th feature set and $\text{sg}(\cdot)$ denotes the stop-gradient operation. This strategy effectively minimizes the distance between diverse domain-specific sub-clusters of the same category, pulling stochastic stylistic fluctuations toward a unified categorical manifold to ensure stable decision responses regardless of environmental shifts.

The joint action of LADA, SAMS, and $\mathcal{L}_\text{con}$ regularizes the latent space into an orderly intra-class inter-domain topology. In this framework, the source feature space provides the structural anchors, while the textual embeddings—containing both domain and category information—act as the shaping force. By explicitly tying visual variations to these textual anchors and minimizing the divergence between branches, TSGST ensures that even domain-dominant features remain clustered within their respective categorical boundaries. This structural refinement ensures that domain-specific variations are logically arranged by the textual guidance rather than being mixed chaotically, effectively alleviating the inter-class confusion prevalent in prior alignment techniques. Consequently, the model maintains distinct inter-class boundaries while accommodating vast intra-class domain shifts, leading to superior generalization in unseen domains.

\subsection{Domain-Aware Query Adapter}

While TSGST constructs a highly structured feature space with a clear intra-class inter-domain topology, the object queries $Q$ in previous query-based architectures often act as domain-blind priors. Although these queries aggregate spatial and visual context through decoding layers, they typically lack the dynamic sensitivity required to exploit the fine-grained manifold established in the latent space. To bridge this gap, we propose the Domain-Aware Query Adapter (DAQA). Acting as an instance-level calibrator, DAQA shifts the perspective from static prior matching to dynamic, domain-aware recalibration. By integrating categorical guidance with representative domain signatures, DAQA ensures that each query "navigates" precisely to its corresponding domain-specific sub-cluster.

\paragraph{Domain Signature as Stylistic Coordinates.}
To inform the queries of their stylistic context, we first derive a robust domain signature $S$. Let $F_{\text{mask}}\in \mathbb{R}^{H\times W\times D}$ denote the mask feature produced by the pixel decoder. As a dense visual representation, $F_{\text{mask}}$ aggregates hierarchical contextual information from the frozen backbone while preserving the spatial layout essential for mask prediction.

Inspired by previous work in style encoding and feature aggregation~\cite{Huang2017Arbitrary,Woo2018Cbam}, we characterize the domain signature by extracting first-order, second-order, and salient statistics directly from $F_{\text{mask}}$. Specifically, we compute the channel-wise mean $\mu(\cdot)$, standard deviation $\sigma(\cdot)$, and maximum values $\max(\cdot)$ over the spatial dimensions. These statistics are concatenated and projected through a multi-layer perceptron (MLP) to yield the domain signature $S\in \mathbb{R}^D$:
\begin{equation}
S = \text{MLP}_{\text{style}}([\mu(F_{\text{mask}}) ; \sigma(F_{\text{mask}}) ; \max(F_{\text{mask}})]).
\end{equation}
In our framework, $S$ provides the necessary context to identify the specific domain region inhabited by the current input on the latent map, providing the baseline for subsequent query modulation.

\paragraph{Domain-Agnostic Semantic Guidance.}
In the decoder pipeline, queries $Q$ enter DAQA after the initial cross-attention layers, having already pre-aggregated certain spatial and visual contexts. To determine the categorical identity of these queries and facilitate precise localization, we extract categorical semantic guidance $E_\text{sem}$ by allowing the queries $Q$ to interact with fixed textual embeddings $\mathcal{E}_{cls}\in \mathbb{R}^{C\times D}$ through a cross-attention mechanism:
\begin{equation}
E_\text{sem} = \text{CrossAttn}(Q, \mathcal{E}_{cls}, \mathcal{E}_{cls}).
\end{equation}
By distilling domain-agnostic knowledge from the stable textual space, $E_\text{sem}$ filters out environmental noise and identifies the categorical center, serving as a reliable semantic reference for the subsequent domain-aware shift.

\paragraph{Domain-Conditioned Recalibration.}
The core of DAQA lies in jointly conditioning the queries on both the categorical guidance $E_\text{sem}$ and the domain signature $S$ to perform a fine-grained refinement. We concatenate $E_\text{sem}$ and $S$ to predict instance-specific modulation parameters via a specialized MLP, denoted as $\text{MLP}_{\text{aff}}$:
\begin{equation}
[\alpha ; \beta] = \text{MLP}_{\text{aff}}([E_\text{sem} ; S]),
\end{equation}
where $\alpha$ and $\beta$ denote the scaling and shifting factors, respectively. These parameters reflect the combined influence of categorical identity and stylistic context. The final domain-refined queries $Q^\prime$ are then generated by applying this modulation as a residual correction to the original queries:
\begin{equation}
Q^\prime = Q+\text{LN}(Q) \odot \alpha + \beta.
\end{equation}
Here, $\odot$ denotes the Hadamard product, and $\text{LN}(\cdot)$ represents Layer Normalization. By retaining the original queries $Q$ through a residual connection, the visual context already aggregated in the queries is preserved. In contrast to previous methods that rely on rigid normalization to force domain-invariance—which often overlooks local cluster variances—our fine-grained recalibration adaptively steers each query toward its corresponding domain-specific sub-cluster. This operation aligns the query's latent position with the actual feature distribution of the current domain, effectively preserving the discriminative semantic details.

In summary, DAQA solidifies the orderly topology of the latent space by transitioning from static prior matching to dynamic localization. By conditioning queries on categorical semantics and domain signature, DAQA ensures that each query and feature "meet" at the most accurate point in the manifold—specifically, at the intra-categorical sub-region corresponding to the current domain.

\subsection{Domain-Aware Decoder Optimizer}
Although DAQA recalibrates object queries $Q^\prime$ to navigate toward domain-specific sub-clusters, a key characteristic of our framework is the use of a universal categorical classifier shared across all domains. While this classifier is jointly optimized during training, it defines its decision boundaries within a unified parameter space for all inputs. As observed in our feature manifold analysis (see \cref{fig:feature_space}), while domain-dominant features remain within their categorical boundaries, the close proximity of sub-clusters from different categories poses a potential risk of inter-class confusion. To resolve the functional tension between domain-specific localization for masking and the requirement for consistent categorical alignment for the shared classifier, we propose the Domain-Aware Decoder Optimizer (DADO).
\paragraph{Decoupled Prediction Heads.}
DADO implements a strategic decoupling of the representation space via branch-specific normalization, ensuring each prediction head operates on a distribution optimized for its respective objective:

1) Categorical Branch (Alignment for Shared Classifier): To ensure the shared linear classifier reliably distinguishes categories across varying environments, DADO employs Adaptive Layer Normalization (AdaLN) to project the domain-aware queries $Q^\prime$ into a unified canonical distribution:
\begin{equation}
Q_{cls}^{\prime \prime} = \text{AdaLN}(Q^\prime, S) = \gamma(S) \odot \text{LN}(Q^\prime) + \delta(S),
\end{equation}
The scaling factor $\gamma$ and shifting factor $\delta$ are dynamically generated from the domain signature S via a lightweight MLP. This rectification compensates for systematic distribution offsets, effectively "pulling" adjacent domain-specific sub-clusters toward a stabilized categorical center to mitigate the risk of inter-class confusion.

2) Mask Branch (Structural Preservation): In contrast, the mask prediction head requires queries to retain their active domain-aware characteristics to accurately interact with the shifted visual features in $F_\text{mask}$. To preserve the localized coordinates established by DAQA, the mask embeddings $Q_{\text{mask}}^{\prime\prime}$ are derived using standard Layer Normalization:
\begin{equation}
Q_\text{mask}^{\prime \prime} = \text{LN}(Q^\prime).
\end{equation}
The final prediction $M$ is generated via the inner product: $M=Q_{\text{mask}}^{\prime\prime}\times F_{\text{mask}}$. This ensures that the spatial segmentation remains strictly anchored to the domain-specific manifestation of objects.

The design of DADO acknowledges that domain-aware navigation and shared-classifier compatibility are divergent functional requirements. If a unified normalization were applied to both branches, the benefits of active domain perception gained in DAQA—essential for masking—would be neutralized by the global alignment required for stable categorization. By decoupling these operations, DADO allows the queries to function simultaneously as domain-sensitive probes for precise localization and domain-invariant representations for robust categorization. This two-stage refinement solidifies the orderly topology of the manifold, ensuring superior generalization in diverse and unseen domains.

\section{Experiments}

\begin{table*}[t] 
    \centering
    \caption{Single-source domain generalization setting. Mean IoU (\%) comparison of different methods. ``--'' indicates the metric is not reported or official source code is unavailable. The (best) and \underline{second best} results are highlighted.}
    \label{tab:g}
    \begin{tabular*}{\textwidth}{@{\extracolsep{\fill}} ll rrrrr rrrrr  @{}}
        \toprule 
        \textbf{Method} & \textbf{Backbone} & \multicolumn{5}{c}{\textbf{Trained on GTAV (G)}} & \multicolumn{5}{c}{\textbf{Trained on Cityscapes (C)}} \\
        \cmidrule(lr){3-7} \cmidrule(lr){8-12}
        & & $\rightarrow$C & $\rightarrow$B & $\rightarrow$M & $\rightarrow$S & Avg.
        & $\rightarrow$B & $\rightarrow$M & $\rightarrow$G & $\rightarrow$S & Avg. \\
        \midrule 
        IBN-Net \cite{Pan2018Two} & ResNet-50 & 33.85 & 32.30 & 37.75 & 27.90 & 32.95 & 48.56 & 57.04 & 45.06 & 26.14 & 44.20 \\
        Iternorm \cite{Huang2019Iterative} & ResNet-50 & 31.81 & 32.70 & 33.88 & 27.07 & 31.37 & 49.23 & 56.26 & 45.73 & 25.98 & 44.30 \\
        RobustNet \cite{Choi2021Robustnet} & ResNet-50 & 36.58 & 35.20 & 40.33 & 28.30 & 35.10 & 50.73 & 58.64 & 45.00 & 26.20 & 45.14 \\
        SHADE \cite{Zhao2022Style} & ResNet-50 & 44.65 & 39.28 & 43.34 & 28.41 & 38.92 & 50.95 & 60.67 & 48.61 & 27.62 & 46.96 \\
        SAN-SAW \cite{Peng2022Semantic} & ResNet-50 & 39.75 & 37.34 & 41.86 & 30.79 & 37.44 & 52.95 & 59.81 & 47.28 & 28.32 & 47.09 \\
        WildNet \cite{Lee2022Wildnet} & ResNet-50 & 44.62 & 38.42 & 46.09 & 31.34 & 40.12 & 50.94 & 58.79 & 47.01 & 27.95 & 46.17 \\
        HRDA \cite{Hoyer2023Domain} & ResNet-101 & 39.63 & 38.69 & 42.21 & -- & -- & -- & -- & -- & -- & -- \\
        TLDR \cite{Kim2023Texture} & ResNet-50 & 46.51 & 42.58 & 46.18 & 36.30 & 42.89 & -- & -- & -- & -- & -- \\
        DPCL \cite{Yang2023Generalized} & ResNet-50 & 44.87 & 40.21 & 46.74 & -- & -- & 52.29 & -- & 46.00 & 26.60 & -- \\
        HGFormer \cite{Ding2023Hgformer} & ResNet-50 & -- & -- & -- & -- & -- & 51.50 & 61.60 & 50.40 & 30.10 & 48.40 \\
        BlindNet \cite{Ahn2024Style} & ResNet-50 & 45.72 & 41.32 & 47.08 & 31.39 & 41.38 & 51.84 & 60.18 & 47.97 & 28.51 & 47.13 \\
        FAMix \cite{Fahes2024simple} & ResNet-50 & 48.15 & \underline{45.61} & 52.11 & 34.23 & 45.03 & \underline{54.07} & 58.72 & 45.12 & 32.67 & 47.65 \\
        DGInStyle \cite{Jia2024Dginstyle} & ResNet-101 & 46.89 & 42.81 & 50.19 & -- & -- & -- & -- & -- & -- & -- \\
        SCSD \cite{Niu2025Exploring} & ResNet-50 & \underline{51.72} & 44.67 & \underline{56.98} & \underline{43.08} & \underline{49.11} & 52.25 & \underline{62.51} & \underline{51.00} & \underline{39.77} & \underline{51.38} \\
        \textbf{LASA (Ours)} & ResNet-50 & \textbf{55.93} & \textbf{50.15} & \textbf{58.56} & \textbf{45.28} & \textbf{52.48} & \textbf{54.97} & \textbf{64.80} & \textbf{52.01} & \textbf{39.82} & \textbf{52.90} \\
        \bottomrule
    \end{tabular*}
\end{table*}

\begin{table}[t] 
    \centering
    \caption{Multi-source setting trained on GTAV + SYNTHIA. Mean IoU (\%) comparison of different methods with ResNet-50 backbone.}
    \label{tab:g_s}
    
    \setlength{\tabcolsep}{1pt}
    \begin{tabular*}{\columnwidth}{@{\extracolsep{\fill}} lrrrr @{}}
        \toprule 
        \textbf{Method} & \multicolumn{4}{c}{\textbf{Trained on G+S}} \\
        \cmidrule(lr){2-5} 
        & $\rightarrow$ C & $\rightarrow$ B & $\rightarrow$ M & Avg. \\
        \midrule 
        IBN-Net \cite{Pan2018Two} & 35.55 & 32.18 & 38.09 & 35.27 \\
        RobustNet \cite{Choi2021Robustnet} & 37.69 & 34.09 & 38.49 & 36.76 \\
        SHADE \cite{Zhao2022Style} & 47.43 & 40.30 & 47.60 & 45.11 \\
        Pin the memory \cite{Kim2022Pin} & 44.51 & 38.07 & 42.70 & 41.76 \\
        AdvStyle \cite{Zhong2022Adversarial} & 39.29 & 39.26 & 41.14 & 39.90 \\
        TLDR \cite{Kim2023Texture} & 48.83 & 42.58 & 47.80 & 46.40 \\
        SPC-Net \cite{Huang2023Style} & 46.36 & 43.18 & 48.23 & 45.92 \\
        FAMix \cite{Fahes2024simple} & 49.41 & \underline{45.51} & 51.61 & 48.84 \\
        SCSD \cite{Niu2025Exploring} & \underline{52.43} & 45.25 & \underline{56.58} & \underline{51.42} \\
        \textbf{LASA (Ours)} & \textbf{55.80} & \textbf{51.71} & \textbf{57.81} & \textbf{55.11} \\
        \bottomrule 
    \end{tabular*}
\end{table}

\begin{table}[t] 
    \centering
    \caption{Adverse condition setting trained on GTAV. Mean IoU (\%) comparison of different methods with ResNet-50 backbone.}
    \label{tab:acdc}
    
    \setlength{\tabcolsep}{2pt}
    \begin{tabular*}{\columnwidth}{@{\extracolsep{\fill}} lrrrrr @{}}
        \toprule 
        \textbf{Method} & \multicolumn{5}{c}{\textbf{Trained on GTAV (G)}} \\
        \cmidrule(lr){2-6} 
        & $\rightarrow$ AN & $\rightarrow$ AS & $\rightarrow$ AR & $\rightarrow$ AF & Avg.\\
        \midrule 
        RobustNet \cite{Choi2021Robustnet} & 6.32 & 29.97 & 33.02 & 32.56 & 25.47 \\
        SHADE \cite{Zhao2022Style} & 8.18 & 30.38 & 35.44 & 36.87 & 27.72 \\
        WildNet \cite{Lee2022Wildnet} & 8.27 & 30.29 & 36.32 & 35.39 & 27.57 \\
        SiamDoGe \cite{Wu2022Siamdoge} & 10.60 & 30.71 & 35.84 & 36.45 & 28.40 \\
        TLDR \cite{Kim2023Texture} & 13.13 & 36.02 & 38.89 & 40.58 & 32.16 \\
        FAMix \cite{Fahes2024simple} & 14.96 & 37.09 & 38.66 & 40.25 & 32.74 \\
        SCSD \cite{Niu2025Exploring} & \underline{15.06} & \underline{41.37} & \underline{42.77} & \underline{43.43} & \underline{35.66} \\
        LASA (Ours) & \textbf{23.70} & \textbf{50.28} & \textbf{43.28} & \textbf{51.28} & \textbf{42.14} \\
        \bottomrule 
    \end{tabular*}
\end{table}

\begin{table}[t]
    \centering
    \caption{Step-by-step ablation study of the LASA framework (mIoU \%). T: TSGST; Con.: Consistency regularization; Q: DAQA; O: DADO.}
    \label{tab:ablation_standard}
    
    \begin{tabular*}{\columnwidth}{@{\extracolsep{\fill}}cccc rrrrr}
        \toprule
        \textbf{T} & \textbf{Con.} & \textbf{Q} & \textbf{O} & \textbf{C} & \textbf{B} & \textbf{M} & \textbf{S} & \textbf{Avg.} \\
        \midrule 
                   &            &            &            & 45.97 & 41.93 & 46.03 & 42.77 & 44.18 \\
        \checkmark &            &            &            & 52.60 & 49.37 & 56.62 & 43.68 & 50.57 \\
        \checkmark & \checkmark &            &            & 55.49 & 49.28 & 58.07 & 45.09 & 51.98 \\
        \checkmark & \checkmark & \checkmark &            & \underline{55.63} & \underline{49.30} & \underline{58.32} & \underline{45.20} & \underline{52.11} \\
        \checkmark & \checkmark & \checkmark & \checkmark & \textbf{55.93} & \textbf{50.15} & \textbf{58.56} & \textbf{45.28} & \textbf{52.48} \\
        \bottomrule
    \end{tabular*}
    
    \vspace{0.3cm} 

    \begin{tabular*}{\columnwidth}{@{\extracolsep{\fill}}cccc rrrrr}
        \toprule
        \textbf{T} & \textbf{Con.} & \textbf{Q} & \textbf{O} & \textbf{AN} & \textbf{AS} & \textbf{AR} & \textbf{AF} & \textbf{Avg.} \\
        \midrule 
                   &            &            &            & 14.12 & 32.55 & 34.37 & 37.17 & 29.55 \\
        \checkmark &            &            &            & 20.03 & 44.02 & 40.59 & 48.55 & 38.30 \\
        \checkmark & \checkmark &            &            & 23.00 & 50.03 & 42.94 & 50.67 & 41.66 \\
        \checkmark & \checkmark & \checkmark &            & \underline{23.38} & \underline{50.13} & \underline{43.23} & \underline{50.84} & \underline{41.90} \\
        \checkmark & \checkmark & \checkmark & \checkmark & \textbf{23.70} & \textbf{50.28} & \textbf{43.28} & \textbf{51.28} & \textbf{42.14} \\
        \bottomrule
    \end{tabular*}
\end{table}

\subsection{Experimental Setup}
\noindent\paragraph{Datasets.}
The evaluation incorporates six semantic segmentation datasets.
Synthetic domains include GTAV (G)~\cite{Richter2016Playing} and SYNTHIA (S)~\cite{Ros2016synthia}.
Real-world domains consist of Cityscapes (C)~\cite{Cordts2016cityscapes}, BDD100K (B)~\cite{Yu2020Bdd100k}, Mapillary (M)~\cite{Neuhold2017mapillary}, and the ACDC (A)~\cite{Sakaridis2021ACDC} dataset which covers four adverse conditions: Fog (AF), Night (AN), Rain (AR), and Snow (AS).
Performance is measured using mIoU over 19 categories, except for SYNTHIA, which uses 16 categories~\cite{Tsai2018Learning}.

\noindent\paragraph{Evaluation Protocols.}
Three zero-shot generalization protocols are employed~\cite{Lee2022Wildnet,Niu2025Exploring}: 
(1) \textit{Single-Source}: Train on G, test on \{C, B, M, S\}.
(2) \textit{Real-to-Mixed}: Train on C, test on \{B, M, G, S\}. 
(3) \textit{Multi-Source}: Train on G+S, test on \{C, B, M\}. 
Robustness is further validated on all ACDC subsets across these settings.

\subsection{Implementation Details}
\noindent\paragraph{Network Architecture.}
The framework uses a ResNet-50 backbone initialized with CLIP pre-trained weights~\cite{Radford2021Learning,Li2023Clip,Fahes2024simple}. Notably, both the CLIP vision backbone and the text encoder are kept frozen throughout all training stages to provide stable semantic priors. 
The segmentation architecture includes a 6-layer multi-scale deformable attention pixel decoder and a 10-layer mask decoder with 100 object queries~\cite{Ding2023Hgformer,Niu2025Exploring}. 
To perform fine-grained query recalibration and global distribution alignment, we deploy the DAQA and DADO modules in layers $[2,3,4,5]$ of the mask decoder.

\noindent\paragraph{Training Strategy.}
Optimization is performed using AdamW~\cite{Loshchilov2017Decoupled} for 225,000 iterations with a batch size of 4 and a learning rate of $1\times10^{-4}$ (PolyLR decay, power 0.9~\cite{Chen2017Rethinking}). 
A decoupled two-phase training scheme is adopted: (1) In the foundation stage (initial 200,000 iterations), the adaptation modules (DAQA and DADO) are frozen, while the pixel decoder, mask decoder, and the text-to-image feature space projector are optimized. (2) In the adaptation stage (remaining 25,000 iterations), we freeze all previously trained segmentation components and exclusively update the parameters of DAQA and DADO.
This staged optimization ensures that the adaptation modules perform alignment based on a finalized feature manifold. During training, images are randomly cropped to $768\times 768$. The total loss comprises weighted cross-entropy (3.0), dice (5.0), mask (5.0), and consistency (10.0) components. All experiments are conducted on 4 $\times$ NVIDIA GeForce RTX 3090 GPUs.

\subsection{Quantitative Analysis}

\paragraph{Synthetic-to-Real Generalization.}
As summarized in~\cref{tab:g}, LASA consistently achieves superior performance in the challenging GTAV $\rightarrow$ Real scenarios.
On the GTAV $\rightarrow$ Cityscapes task, LASA reaches an mIoU of 55.93\%, outperforming the recent SOTA method SCSD by a significant margin of $+$4.21\%.
Across all target domains, LASA achieves an average mIoU of 52.48\%, which is $+$3.37\% higher than SCSD.
These substantial gains demonstrate that by stabilizing the feature manifold with structural anchors and fine-grained VLM guidance, LASA effectively addresses manifold distortion.
This allows the model to preserve the discriminative semantic nuances that are typically suppressed by rigid designs or coarse-grained randomization.

\paragraph{Real-to-Mixed Generalization.}
When trained on the real-world Cityscapes dataset, LASA continues to lead the competition with an average mIoU of 52.90\%.
It surpasses all ResNet-50-based methods, confirming its ability to adapt to mixed environmental variations while maintaining high categorical discriminability.

\paragraph{Multi-Source Generalization Results.}
\cref{tab:g_s} reports the GTAV + SYNTHIA $\rightarrow$ Real results. LASA achieves 55.11\% mIoU, significantly surpassing SCSD (51.42\%). This performance confirms LASA's efficacy in synergistically aggregating multi-source knowledge without distribution interference.

\paragraph{Robustness in Adverse Scenarios.}
To further stress-test the model's adaptability in extreme environments, we evaluate LASA on the ACDC dataset.
As summarized in \cref{tab:acdc}, LASA sets a new state-of-the-art for overall robustness with an average mIoU of 42.14\%, representing a significant $+$6.48\% increase over the previous methods.
The superiority of our framework is most pronounced in the most challenging scenarios: in low-illumination Night (AN) environments where visual cues are severely degraded, LASA improves the mIoU from 15.06\% (SCSD) to 23.70\% (a remarkable $+$8.64\% increase). In visibility-impaired Snow (AS) conditions, it reaches 50.28\% mIoU, outperforming SCSD (41.37\%) by +8.91\%.

\paragraph{Manifold Quality Analysis.}
\begin{table}[t]
\centering
\caption{Quantitative manifold quality comparison between the proposed approach and recent methods.}
\label{tab:manifold_quality}
\begin{tabular*}{\columnwidth}{@{\extracolsep{\fill}}lrrr}
\toprule
\textbf{Metric} & \textbf{SCSD}~\cite{Niu2025Exploring} & \textbf{LASA (Ours)} & \textbf{$\Delta$} \\
\midrule
Silhouette~\cite{Rousseeuw1987Silhouettes} $\uparrow$ & 0.4264 & \textbf{0.6575} & {$+$0.2312} \\
DBI~\cite{Davies1979cluster} $\downarrow$ & 0.9073 & \textbf{0.4524} & {$-$0.4549} \\
IC-PAD~\cite{BenDavid2010theory} $\uparrow$ & $-$0.5015 & \textbf{0.3922} & {$+$0.8937} \\
Fisher's $J$~\cite{Fisher1936use,Wen2016discriminative} $\uparrow$ & 8.0081 & \textbf{50.8793} & {$+$42.8712} \\
\bottomrule
\end{tabular*}
\end{table}
To quantitatively validate the structural integrity of the latent space, we evaluate the feature manifold using high-dimensional mask features extracted from the decoder, with four literature-backed metrics summarized in \cref{tab:manifold_quality}.
The Silhouette Score~\cite{Rousseeuw1987Silhouettes} ($[-1,1]$, $\uparrow$) measures intra-class compactness versus inter-class separation; the Davies-Bouldin Index (DBI)~\cite{Davies1979cluster} ($[0,+\infty)$, $\downarrow$) quantifies average similarity to the most confusable neighbor.
LASA achieves a Silhouette Score of 0.66 ($+$0.23) and reduces DBI from 0.91 to 0.45, confirming tighter clusters and sharper boundaries.
The Intra-Class Proxy $\mathcal{A}$-distance (IC-PAD)~\cite{BenDavid2010theory} ($[-1,1]$, $\uparrow$) adapts domain divergence theory to an intra-class setting by applying the Proxy $\mathcal{A}$-distance within each semantic class independently and reporting the mean score across all classes; a higher score indicates well-organized intra-class domain structure rather than an undifferentiated collapse.
The prior state-of-the-art yields IC-PAD of $-$0.50, confirming that coarse-grained randomization destroys intra-class domain structure; LASA raises this to $+$0.39 ($+$0.89).
Fisher's $J$~\cite{Fisher1936use,Wen2016discriminative} ($[0,+\infty)$, $\uparrow$) jointly captures within-class scatter and between-class separation, improving from 8.01 to 50.88.
These findings are corroborated by the t-SNE visualizations in \cref{fig:feature_space}, where LASA exhibits well-separated class clusters with visible domain sub-structure, in contrast to the inter-class overlap and domain-collapsed distributions of prior methods.

\subsection{Qualitative Analysis}

\begin{figure}
    \includegraphics[width=\linewidth]{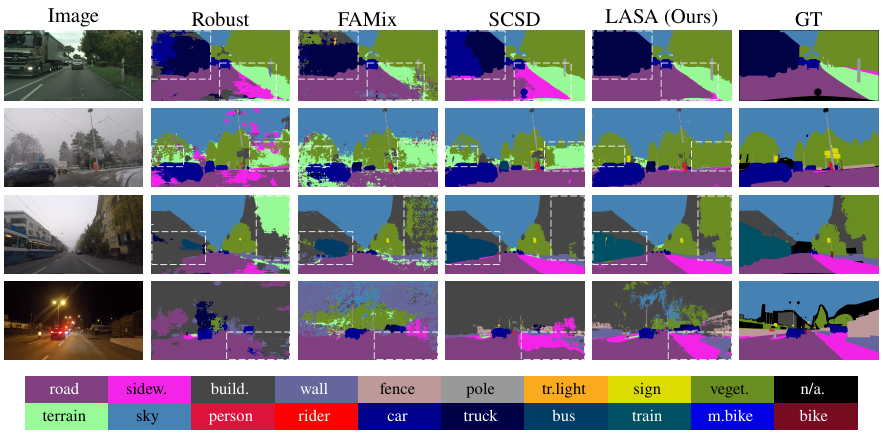}
    \caption{Visualization comparison between the proposed approach and recent methods.}
    \label{fig:qualitative}
\end{figure}

To qualitatively assess the effectiveness of LASA, we present visualization results across diverse target domains, spanning standard urban scenes and extreme adverse environments.

LASA shows greater robustness in visibility-impaired environments, such as nighttime and fog. While previous methods suffer from categorical confusion caused by manifold distortion, LASA maintains clear object boundaries and consistent categorical labels.


\subsection{Ablation Study}
Using GTAV as the single source domain, we conduct an incremental ablation study to evaluate the contribution of each LASA component. The results are summarized in \cref{tab:ablation_standard}.

\paragraph{Text-and-Source-Guided Style Transfer.}
As the foundation of our framework, the TSGST module yields the most substantial performance gain. The single-path configuration—incorporating Language-Anchored Distribution Alignment (LADA) and Source-Anchored Manifold Stabilization (SAMS)—improves the average mIoU from 44.18\% to 50.57\% (+6.39\%). This significant improvement demonstrates that providing fine-grained guidance via textual priors while maintaining structural anchors effectively prevents representation collapse and addresses manifold distortion.

\paragraph{Dual-path Alignment.}
Building upon the stabilization provided by the TSGST module, the dual-path mechanism further enforces semantic consistency via $\mathcal{L}_\text{con}$. This configuration increases the average mIoU to 51.98\% and notably lifts performance on the adverse ACDC dataset from 38.30\% to 41.66\%. These results confirm that probabilistic consistency regularization effectively filters stylistic noise, forcing the model to focus on domain-invariant semantic cues rather than transient environmental characteristics.

\paragraph{Query-level Calibration and Rectification.}
Final performance refinements are provided by the adaptation modules. The DAQA module recalibrates object queries using domain-aware signatures to restore suppressed discriminative details, reaching 52.11\% mIoU. Finally, the complete LASA framework, incorporating the DADO global distribution rectifier, achieves a peak performance of 52.48\%. Their synergy is most pronounced in extreme conditions, pushing the ACDC average to 42.14\%, as they ensure the resulting query distributions remain aligned with the shared classifier for consistent categorical responses.

\section{Conclusion}
In this paper, we presented Language-and-Source-Anchored Alignment (LASA), a novel framework for Domain Generalization Semantic Segmentation (DGSS). LASA is specifically designed to address manifold distortion and restore suppressed semantic details to ensure robust generalization. By introducing the Text-and-Source Guided Style Transfer (TSGST), we effectively utilize fine-grained VLM guidance and structural anchors to construct a coherent and stabilized feature manifold. To further refine this space, the Domain-Aware Query Adapter (DAQA) recalibrates object queries to restore suppressed discriminative details, while the Domain-Aware Decoder Optimizer (DADO) aligns the resulting query distributions with a shared classifier to ensure consistent categorical responses. Extensive experiments across diverse benchmarks demonstrate that LASA consistently achieves state-of-the-art performance, with particularly remarkable gains under extreme conditions such as nighttime and foggy scenarios.

\begin{acks}
This work was supported by the New Generation Artificial Intellig-ence-National Science and Technology Major Project (No.\ 2025ZD\allowbreak0122703), the National Science and Technology Major Project (No. 2025YFE01\allowbreak13500), the National Science Fund for Distinguished Young Scholars (No. 62525605), the National Natural Science Foundation of China (No. U25B2066, No. U22B2051, and No. 62272401), and the Xiamen Municipal Science and Technology Bureau, China (3502ZC-QXT2024009).
\end{acks}

\bibliographystyle{ACM-Reference-Format}
\balance
\bibliography{mm2026}


\end{document}